\documentclass[conference]{IEEEtran}
\IEEEoverridecommandlockouts
\usepackage{pifont}
\usepackage{cite}
\usepackage{amsmath,amssymb,amsfonts}
\usepackage{algorithmic}
\usepackage{graphicx}
\usepackage{textcomp}
\usepackage[table,xcdraw]{xcolor} 
\def\BibTeX{{\rm B\kern-.05em{\sc i\kern-.025em b}\kern-.08em
    T\kern-.1667em\lower.7ex\hbox{E}\kern-.125emX}}

\usepackage{hyperref}

\begin{document}

\title{Training-Free Point Cloud Recognition Based on Geometric and Semantic Information Fusion
\thanks{$^*$Equal contribution.\hspace{1em}$^\dag$Corresponding author.\\This work is supported by National Key Research and Development Program of China (2022YFB3303101).}
}

\author{\IEEEauthorblockN{Yan Chen$^{*}$}
\IEEEauthorblockA{\textit{Institute of Data and Information} \\
\textit{Tsinghua University}\\
Shenzhen, China \\
chenyan23@mails.tsinghua.edu.cn}
\and
\IEEEauthorblockN{Di Huang$^{*,\dag}$}
\IEEEauthorblockA{\textit{Institute of Data and Information} \\
\textit{Tsinghua University}\\
Shenzhen, China \\
dihuangdylan@gmail.com}
\and
\IEEEauthorblockN{Zhichao Liao}
\IEEEauthorblockA{\textit{Institute of Data and Information} \\
\textit{Tsinghua University}\\
Shenzhen, China \\
liaozc23@mails.tsinghua.edu.cn}
\and
\IEEEauthorblockN{Xi Cheng}
\IEEEauthorblockA{\textit{Institute of Data and Information} \\
\textit{Tsinghua University}\\
Shenzhen, China \\
chengxi23@tsinghua.org.cn}
\and
\IEEEauthorblockN{Xinghui Li}
\IEEEauthorblockA{\textit{Intelligent Creation} \\
\textit{ByteDance}\\
Beijing, China \\
li-xh21@tsinghua.org.cn}
\and
\IEEEauthorblockN{Long Zeng$^{\dag}$}
\IEEEauthorblockA{\textit{Institute of Data and Information} \\
\textit{Tsinghua University}\\
Shenzhen, China \\
zenglong@sz.tsinghua.edu.cn}
}

\maketitle

\begin{abstract}
The trend of employing training-free methods for point cloud recognition is becoming increasingly popular due to its significant reduction in computational resources and time costs. However, existing approaches are limited as they typically extract either geometric or semantic features. To address this limitation, we are the first to propose a novel training-free method that integrates both geometric and semantic features. For the geometric branch, we adopt a non-parametric strategy to extract geometric features. In the semantic branch, we leverage a model aligned with text features to obtain semantic features. Additionally, we introduce the GFE module to complement the geometric information of point clouds and the MFF module to improve performance in few-shot settings. Experimental results demonstrate that our method outperforms existing state-of-the-art training-free approaches on mainstream benchmark datasets, including ModelNet and ScanObiectNN.
\end{abstract}

\begin{IEEEkeywords}
Point Cloud, Training-free, Feature Fusion, Few-shot Classification
\end{IEEEkeywords}

\section{Introduction}
\label{sec:intro}

Point cloud recognition \cite{guo2020deeplearning3dpoint,xiao2023unsupervised} has seen significant advancements nowadays, driven by the growing applications \cite{liu2019deep,liao2024freehand,wang2024detdiffusion,che2019object,wang2019dynamic} in various fields such as autonomous driving, robotics, and augmented reality. Previous methods \cite{qi2017pointnet,qi2017pointnet++,zhao2021point,cheng2024constraint,qian2022pointnext,lin2021learning} for point cloud recognition typically rely on large computational resources and extensive time costs for model training, making them less feasible for real-time or resource-constrained applications. These limitations necessitate the exploration of more efficient approaches.
The advent of training-free methods offers a promising solution to these challenges. Unlike previous methods, training-free approaches do not require learning model parameters, thereby significantly reducing the costs. For instance, \cite{zhang2023parameterneedstartingnonparametric} achieves classification outcomes by using a memory bank to match point cloud features. \cite{xue2023uliplearningunifiedrepresentation, xue2024ulip2scalablemultimodalpretraining} leverage contrastive learning to align text features with 3D or image features of point clouds, enabling zero-shot classification. However, these training-free methods focus on either geometric features or semantic features exclusively. This limits the model's ability to  represent and understand point clouds comprehensively. 

The importance of feature fusion in the point cloud domain cannot be overstated. For example, combining local and global geometric features has been shown to achieve higher performance in point cloud recognition tasks\cite{qi2017pointnet++,sheshappanavar2020novel,sheshappanavar2021patchaugment,anvekar2023gpr,sheshappanavar2021dynamic}.
Fusion operations combine the strengths of different feature representations, enhancing the overall performance of the model. Inspired by these successful fusion techniques\cite{pointgs,tran2021feature,sun2005new,zhang2018exfuse},
our work highlights the importance of integrating both semantic and geometric features for point cloud recognition. This fusion is crucial as it enables a more comprehensive understanding of the point cloud data by capturing both the shape and the meaning of the objects within the point cloud. To the best of our knowledge, this paper is the first to combine semantic and geometric features for point cloud recognition in a training-free context.


Our approach requires constructing a feature memory (MEM), which is a collection of features obtained from the support set using an IF-encoder. The IF-encoder is a feature extractor composed of a geometric encoder and a semantic encoder. The geometric encoder employs non-parametric strategies such as farthest point sampling (FPS), k-nearest neighbors (k-NN), and pooling operations to extract features. The semantic encoder uses a pre-trained model that aligns point clouds with data containing semantic information to extract features. For the few-shot classification task, we design a clustering method to select representative samples. We then calculate the cosine similarity between the extracted features from the test point clouds and those stored in the feature memory (MEM) for feature matching, and compare the matching results with the one-hot encoded category labels to obtain the classification results. Finally, we ensemble our classifier with ULIP's zero-shot classifier to further improve our classification 

\begin{figure*}[t]
    \centering
    \includegraphics[width=1\linewidth]{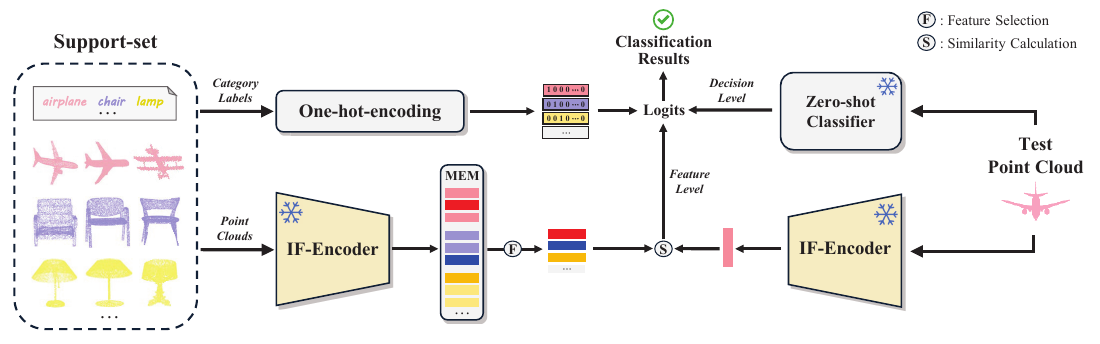}
    \caption{Framework of our training-free information fusion approach for point cloud recognition.}
    \label{fig:framework}
\end{figure*}

\noindent accuracy. Experimental results demonstrate that our training-free method achieves state-of-the-art performance across several benchmark datasets, including ModelNet \cite{wu20153d} and ScanObjectNN \cite{uy2019revisiting}.


\section{Related Work}
\label{sec:related}

Our work is related to training-free models, few-shot learning, and information fusion methods for point cloud recognition. We explore them in this section.

\noindent{\bfseries Training-free models.} Recent advancements in training-free models for 3D point cloud recognition can be categorized into zero-shot and memory-based approaches. Zero-shot models, such as PointCLIP\cite{zhang2021pointclippointcloudunderstanding}, PointCLIP-V2\cite{zhu2023pointclipv2promptingclip}, CLIP2Point\cite{huang2023clip2pointtransferclippoint}, ViT-Lens\cite{lei2024vitlensinitiatingomnimodalexploration}, ULIP\cite{xue2023uliplearningunifiedrepresentation}, and ULIP-2\cite{xue2024ulip2scalablemultimodalpretraining}, prealign natural language with 2D or 3D data, enabling direct use of learned semantic features for classification without additional training. For instance, PointCLIP\cite{zhang2021pointclippointcloudunderstanding} converts 3D clouds into 2D images for CLIP\cite{radford2021learning} processing, while ULIP-2\cite{xue2024ulip2scalablemultimodalpretraining} introduces a scalable tri-modal framework that enhances zero-shot classification by automatically generating language descriptions for 3D shapes. Memory-based models, like Point-NN\cite{zhang2023parameterneedstartingnonparametric}, TIP-Adapter\cite{zhang2021tip}, and PointTFA\cite{pointtfa}, use stored features in a key-value database to match and classify new data without additional training. PointTFA, for example, refines query clouds using a representative memory cache, achieving performance close to fine-tuned models. These approaches highlight the effectiveness of using pre-existing knowledge and memory mechanisms in 3D point cloud recognition.

\noindent{\bfseries Few-shot learning.} Few-shot learning for 3D point cloud recognition typically involves leveraging a pre-trained base model, where an adapter module is inserted and fine-tuned with limited new data. This approach allows the model to retain the knowledge from the base model while adapting to new tasks with fewer training resources. For instance, methods like PointCLIP\cite{zhang2021pointclippointcloudunderstanding} and CLIP2Point\cite{huang2023clip2pointtransferclippoint} use learnable adapters that fine-tune specific layers or units to better align with the downstream tasks. PointCLIP\cite{zhang2021pointclippointcloudunderstanding} adjusts 2D depth features with global information, while CLIP2Point\cite{huang2023clip2pointtransferclippoint} fine-tunes the gate unit in feature fusion. Additionally, models like PointTFA\cite{pointtfa} support few-shot learning by selecting a small set of representative samples from each class to construct a memory bank. During testing, query point clouds are matched with this memory without requiring any fine-tuning of network parameters, enabling effective classification in a few-shot setting.

\noindent{\bfseries Information fusion.} Recent advancements in information fusion methods for 3D point cloud recognition have significantly improved the accuracy and robustness of various tasks such as classification, segmentation, and retrieval. Previous methods often rely on geometric information alone, as seen in works like the GBNet\cite{qiu2021geometricbackprojectionnetworkpoint}, or emphasize the importance of semantic spaces, as demonstrated by the PointGS framework\cite{pointgs}. Other approaches, such as GRADE\cite{li2024graph}, incorporate self-learning strategies to better bridge domain gaps by directly delivering semantic information. Meanwhile, methods like GSLCN\cite{liang2023long} focus on constructing optimal graph structures to enhance feature extraction from both short and long-range dependencies. These fusion techniques typically enhance point cloud model performance but often overlook additional contextual information. Our approach, the first to combine natural language semantic descriptions with 3D point cloud data in a training-free context, enriches 3D model representation by integrating external semantic context with existing geometric and semantic fusion methods.



\section{Approach}
Fig.\ref{fig:framework} shows the framework of our training-free point cloud recognition approach based on geometric and semantic information fusion. Inspired by \cite{pointgs,qiu2021geometricbackprojectionnetworkpoint,zhang2023parameterneedstartingnonparametric}, our geometric feature extraction branch adopts non-parametric strategies and incorporates spherical coordinate relation and edge information. For the semantic feature extraction branch, the best backbone, the 3D encoder from ULIP\cite{xue2023uliplearningunifiedrepresentation}, is chosen. For the fusion of geometric and semantic features, we select the method of weighted summation, introducing a hyperparameter \(\alpha\). To enhance the performance of our approach on few-shot learning tasks, we propose the Memory Feature Filtering (MFF) method to extract the most representative samples from each category. Additionally, we propose the Geometric Feature Enhancement (GFE) method to enhance the geometric information of point clouds.



\begin{figure}[t]
    \flushleft
    \includegraphics[width=1\linewidth]{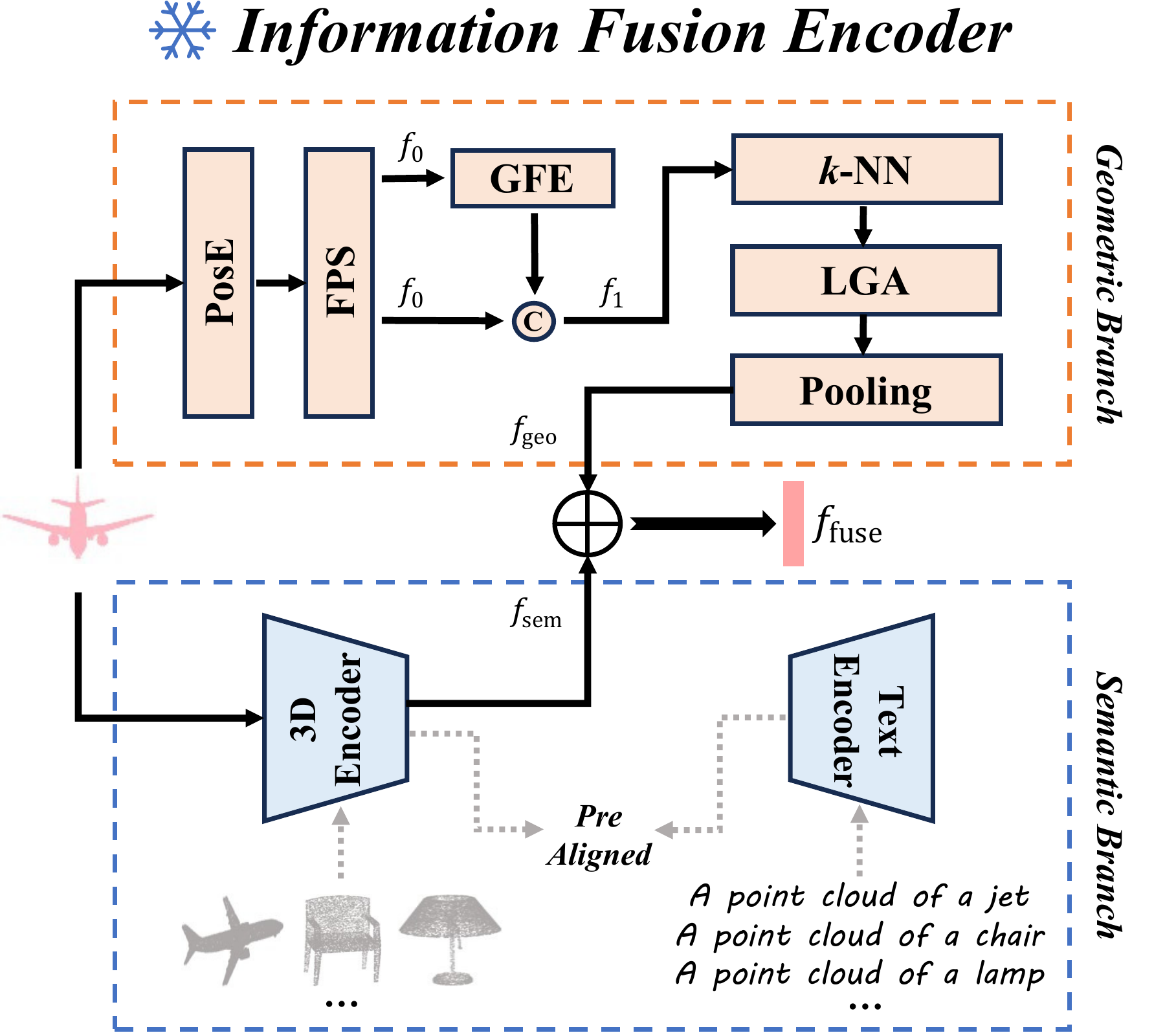}
    \caption{Our information fusion encoder.}
    \label{fig:ifencoder}
\end{figure}

\subsection{Geometric and Semantic Information Fusion}
As shown in Fig.\ref{fig:ifencoder}, we propose a dual-branch framework for feature extraction from 3D point cloud data, integrating both geometric and semantic features. The method leverages a combination of training-free geometric processing techniques and a pre-trained 3D encoder to capture a comprehensive representation of the 3D data. 

\noindent{\bfseries Geometric Feature Extraction.} The geometric feature extraction branch begins by applying a position encoding (PosE) to the input point cloud $\mathit{p}$, followed by a farthest point sampling (FPS) operation to downsample the points. The resulting features are further refined by concatenating the initial features with the output of a geometric feature enhancement (GFE) module, and then processed through a \textit{k}-nearest neighbors (\textit{k}-NN) operation to capture local neighborhood information. Finally, these features are passed through a local geometric aggregation (LGA) layer and pooled to produce the final geometric feature representation.

\noindent{\bfseries Semantic Feature Extraction.} In parallel to the geometric feature extraction, the semantic features are directly extracted from the point cloud using a pre-trained ULIP 3D encoder, which has been aligned with natural language descriptions during the pretraining phase. 

\noindent{\bfseries Feature Fusion.} The final step involves combining the geometric and semantic features to form a unified feature representation. This is achieved by computing a weighted sum of the two sets of features. This fused feature $\mathit{f}_{\text{fuse}}$ effectively encapsulates both the local geometric structures and the high-level semantic information, making it suitable for downstream classification tasks.

The entire process can be mathematically described as:
\begin{equation}
\mathit{f}_0 = \text{FPS}\left(\text{PosE}\left(\mathit{p}\right)\right),
\end{equation}
\begin{equation}
\mathit{f}_1 = \text{Concat}\left(\mathit{f}_0, \text{GFE}\left(\mathit{f}_0\right)\right),
\end{equation}
\begin{equation}
\mathit{f}_{\text{geo}} = \text{Pooling}\left(\text{LGA}\left(\textit{k}\text{-NN}\left(\mathit{f}_1\right)\right)\right),
\end{equation}
\begin{equation}
\mathit{f}_{\text{sem}} = \text{ULIP}\left(\mathit{p}\right),
\end{equation}
\begin{equation}
\mathit{f}_{\text{fuse}} = \alpha \cdot \mathit{f}_{\text{geo}} + \left(1 - \alpha\right) \cdot \mathit{f}_{\text{sem}} \ .
\end{equation}

\subsection{Memory Feature Filtering}







In our feature database, categories often contain features with shared characteristics. To enhance memory efficiency and model performance, it is crucial to select key features that best capture the essence of each category. This selection process named Memory Feature Filtering (MFF) is particularly beneficial in a $K$-shot setting, as choosing $K$ such features can significantly reduce redundancy while preserving essential characteristics for downstream tasks.

We achieve MFF by implementing the \textit{K-Means\texttt{++}} clustering algorithm, which improves the initial centroid selection by considering the data point distribution. This approach mitigates issues from random initialization and accelerates convergence. Given a set of feature vectors $\mathit{F}_i$ for the category $i$, we apply MFF to partition these vectors into $K$ clusters. The centroids from all categories are then aggregated to form the key feature set $\mathit{F}_{\text{key}}$, expressed as:

\begin{equation}
    \mathit{F}_{\text{key}} = \bigcup_{i=1}^{N} \left\{ \text{MFF}\left(\mathit{F}_i, K\right) \right\},
\end{equation}
where $N$ denotes the total number of categories.

This approach reduces the overall computational load while ensuring that the memory retains essential information, vital for efficient downstream processing.

\subsection{Geometric Feature Enhancement}
To improve the geometric representation of 3D point clouds, we integrate additional features beyond the standard 3D coordinates \((x, y, z)\), based on methodologies from two key references. First, inspired by \cite{pointgs}, we enhance the feature set by converting Cartesian coordinates into a spherical coordinate system, yielding three pairs of angles \((\theta_X, \phi_X)\), \((\theta_Y, \phi_Y)\), \((\theta_Z, \phi_Z)\), which provide supplementary spatial information. Second, following \cite{qiu2021geometricbackprojectionnetworkpoint}, we extend the point representation  by incorporating edge vectors ($e=f_{j} - f_{i}$) and their lengths ($l=|f_{j} - f_{i}|$) from the two nearest neighbors $f_{j1}$, $f_{j2}$ of each point \(f_{i}\). The cross product of these vectors gives a normal vector ($nv = e_1 \times e_2$), further enriching the geometric relations within the point cloud. The entire process can be
mathematically described as: 
\begin{equation}
\text{GFE}(f) = \left(\theta_X,\phi_X,\theta_Y,\phi_Y,\theta_Z,\phi_Z,\textit{nv},e_1,e_2,l_1,l_2\right),
\end{equation}

\begin{table*}[htbp]
\caption{Performance comparison based on classification accuracy (\%).}
\centering
\resizebox{\textwidth}{!}{
\begin{tabular}{ccccccccc}

\hline
Approach     & Settings & Feature type & 3D Encoder                & ModelNet10 & ModelNet40 & OBJ\_ONLY & OBJ\_BG & OBJ\_T50RS \\ \hline
ULIP-1 \cite{xue2023uliplearningunifiedrepresentation}       & 0-shot$^{a}$   & sem          & \ding{52} & -          & 60.40      & -         & 48.05   & -          \\
ULIP-2 \cite{xue2024ulip2scalablemultimodalpretraining}       & 0-shot$^{a}$   & sem          & \ding{52} & -          & 75.60      & -         & -       & -          \\
RECON \cite{qi2023contrast}        & 0-shot$^{a}$   & sem          & \ding{52} & 75.60      & 61.70      & 43.70     & 40.40   & 30.50      \\
OpenShape \cite{liu2024openshape}    & 0-shot$^{a}$   & sem          & \ding{52} & -          & 85.30      & -         & 56.70   & -          \\
VIT-Lens \cite{lei2024vit}     & 0-shot$^{a}$   & sem          & \ding{56}                  & -          & 87.60      & -         & 60.10   & -          \\
PointCLIP \cite{zhang2021pointclippointcloudunderstanding}   & 16-shot$^{b}$  & sem          & \ding{56}                  & -          & 87.20      & -         & -       & -          \\
PointCLIP-V2 \cite{zhu2023pointclipv2promptingclip} & 16-shot$^{b}$  & sem          & \ding{56}                 & -          & 89.55      & -         & -       & -          \\
CLIP2Point \cite{huang2023clip2pointtransferclippoint}  & 16-shot$^{b}$  & sem          & \ding{56}                  & -          & 87.46      & -         & -       & -          \\
\rowcolor[gray]{0.9} PointTFA \cite{pointtfa}    & 16-shot$^{a}$  & sem          & \ding{52} & 92.62      & 89.79      & 80.90     & 82.10   & 67.18      \\
PointNN \cite{zhang2023parameterneedstartingnonparametric}      & full-shot$^{a}$ & geo          & \ding{56}                  & -          & 81.80      & 74.90     & 71.10   & 64.90      \\
\rowcolor[gray]{0.7} PointTFA \cite{pointtfa}    & full-shot$^{a}$ & sem          & \ding{52} & 93.17      & 90.88      & 83.48     & 84.85   & 68.22      \\                \hline
\rowcolor[gray]{0.9} \textbf{Ours}  & 16-shot$^{a}$  & geo+sem      & \ding{52} & \underline{\textbf{93.06}} & \underline{\textbf{90.48}} & \underline{\textbf{83.30}} & \underline{\textbf{83.48}} & \underline{\textbf{75.12}} \\
\rowcolor[gray]{0.7 } \textbf{Ours}  & full-shot$^{a}$ & geo+sem      & \ding{52} & \underline{\textbf{93.61}} & \underline{\textbf{92.10}} & \underline{\textbf{85.03}} & \underline{\textbf{85.37}} & \underline{\textbf{77.38}} \\ \hline
\end{tabular}

}

\parbox{\textwidth}{
\scriptsize\raggedright\hspace{1em}X-shot$^{a}$: Training-free.\ \ X-shot$^{b}$: Fine-tune.
}

\vspace{-2em}
\label{tab:comparison}
\end{table*}


\begin{table}[htbp]
\caption{Ablation experiment on the ModelNet40 dataset.}
\centering 
\setlength{\tabcolsep}{8pt} 
\begin{tabular}{ccccc}
\hline
SemEnc       & GeoEnc       & GFE          & MFF          & Accuracy    \\ \hline
            & \ding{52}   & \ding{52}   & \ding{52}   &     72.49     \\
\ding{52}   &             & \ding{52}   & \ding{52}   &     88.65        \\
\ding{52}   & \ding{52}   &             & \ding{52}   &     89.75        \\
\ding{52}   & \ding{52}   & \ding{52}   &             &     86.47      \\
\ding{52}   & \ding{52}   & \ding{52}   & \ding{52}   &     \underline{\textbf{90.48}}        \\ \hline
\end{tabular}
\vspace{-1em}
\label{tab:ablation}
\end{table}
\noindent These enhanced features provide richer spatial information for our model, improving performance in classification tasks.

\section{Experiments}
\label{sec:exp}

\subsection{Experimental Settings}
\noindent{\bfseries Dataset.} We demonstrate our performance on three benchmarks: ModelNet10\cite{wu20153d}, ModelNet40 \cite{wu20153d}, and ScanObjectNN \cite{uy2019revisiting}. ModelNet10\cite{wu20153d} and ModelNet40\cite{wu20153d} are widely used benchmark datasets for 3D shape classification. ModelNet10\cite{wu20153d} consists of 4,899 3D object models across 10 categories, while ModelNet40\cite{wu20153d} contains 12,311 3D object models spanning 40 categories. These datasets provide standardized training and test sets and are typically used to evaluate the generalization ability of 3D shape classification algorithms. ScanObjectNN\cite{uy2019revisiting} is a more complex 3D point cloud dataset containing objects scanned from the real world. Unlike the ModelNet datasets, the point clouds in ScanObjectNN\cite{uy2019revisiting} are noisier and exhibit more variations in pose, making it an effective benchmark for assessing an algorithm's robustness in real-world scenarios. ScanObjectNN\cite{uy2019revisiting} includes three variants: OBJ\_ONLY, OBJ\_BG, and OBJ\_T50RS, which are used to test the robustness and adaptability of algorithms.

We adopt the whole training set to construct our feature memory, and the full test set for evaluation. Notably, our architecture requires no training during the entire classification process, so it only takes a few minutes to obtain classification results. To ensure fairness in comparison, we employ the modelnet40\_normal\_resampled version of the ModelNet40 dataset, consistent with the version used by PointTFA\cite{pointtfa}.

\subsection{Comparative Experiments}
We compare our approach with existing training-free and fine-tuned methods. As shown in Table \ref{tab:comparison}, our approach shows state-of-the-art performance in both full-shot and 16-shot settings on three benchmark datasets: ModelNet10\cite{wu20153d}, ModelNet40\cite{wu20153d}, and ScanObjectNN\cite{uy2019revisiting}. This indicates that our approach is not only effective on synthetic data but also maintains high accuracy and stability in challenging real-world scenarios. Furthermore, our approach can achieve accurate classification results under limited resources, relying solely on feature memory without training.

\subsection{Ablation Study}

For the 16-shot classification task, we conduct ablation experiments. We first investigate the performance of our approach when only geometric information or only semantic information is used. Additionally, we examine the impact of the GFE and MFF modules on the approach's performance, with the results shown in Table \ref{tab:ablation}. The ablation study results indicate that both geometric and semantic information contribute significantly to the approach's performance. When using these features independently, our approach already achieves satisfactory accuracy; however, combining them leads to a substantial improvement, demonstrating the effectiveness of the geometric-semantic fusion strategy. Moreover, incorporating the GFE and MFF modules further enhances the approach's accuracy. In particular, the MFF module contributes to a 4.01\% increase in accuracy compared to random sample extraction, providing important insights for sample selection in few-shot learning. The experiments demonstrate that these components play a crucial role in extracting and integrating multi-level features.

\section{Conclusion}
\label{sec:conclusion}
This work, for the first time, presents a novel approach to point cloud recognition that integrates both geometric and semantic information in a training-free framework. The efficiency and understanding of point clouds are improved by selecting key features from the feature memory and enriching geometric feature extraction. Extensive evaluations on benchmark datasets such as ModelNet and ScanObjectNN demonstrate that our method achieves state-of-the-art performance, showing the effectiveness of our feature fusion strategy for accurate point cloud recognition without training.

\clearpage
\bibliographystyle{IEEEbib}
\bibliography{main}

\end{document}